\documentclass[acmtog,nonacm,screen]{acmart}

\usepackage[utf8]{inputenc} 
\usepackage[T1]{fontenc}    
\usepackage{url}            
\usepackage{balance}

\usepackage{graphicx}
\usepackage{subcaption}
\usepackage{tikz}
\usepackage{pgfplots}
\pgfplotsset{compat=1.18} 
\usepackage{xcolor}         

\usepackage{tabulary}
\usepackage{pgfplotstable}
\usepackage{multirow}
\usepackage[para,online,flushleft]{threeparttable}
\usepackage[labelfont=bf,textfont=it,font=small]{caption}

\usepackage{parskip}
\usepackage{microtype}      
\usepackage{enumitem}

\graphicspath{{Figures/}}

\hypersetup{colorlinks,
  linkcolor=blue,
  citecolor=blue,
  urlcolor=blue}

\newcommand{\cref}[2]{\hyperref[#2]{#1~\ref*{#2}}}
\newcommand{\colref}[2]{\hyperref[#2]{#1~\ref*{#2}}}

\newcommand{\figref}[1]{\colref{Fig.}{#1}}

\newcommand{\secref}[1]{\colref{Section}{#1}}

\newcommand{\coloredref}[2]{\hyperref[#2]{#1~\ref*{#2}}}
\newcommand{\coloredsubref}[3]{\hyperref[#2]{#1~\ref*{#2}{#3}}}
\newcommand{\Algref}[1]{\hyperref[#1]{Algorithm~\ref*{#1}}}

\graphicspath{{Figures/}}
\DeclareGraphicsExtensions{.pdf,.png,.jpg}

\begin{document}

\title{Latent Diffusion Models for Structural Component Design}

\author{Ethan Herron}
\email{edherron@iastate.edu} 
\affiliation{%
  \institution{Iowa State University}
  \city{Ames}
  \state{Iowa}
  \country{USA}
}

\author{Jaydeep Rade}
\email{jrrade@iastate.edu} 
\affiliation{%
  \institution{Iowa State University}
  \city{Ames}
  \state{Iowa}
  \country{USA}
}

\author{Anushrut Jignasu}
\email{ajignasu@iastate.edu} 
\affiliation{%
  \institution{Iowa State University}
  \city{Ames}
  \state{Iowa}
  \country{USA}
}

\author{Baskar Ganapathysubramanian}
\email{baskarg@iastate.edu}
\affiliation{%
  \institution{Iowa State University}
  \city{Ames}
  \state{Iowa}
  \country{USA}
}

\author{Aditya Balu}
\email{baditya@iastate.edu} 
\affiliation{%
  \institution{Iowa State University}
  \city{Ames}
  \state{Iowa}
  \country{USA}
}

\author{Soumik Sarkar}
\email{soumiks@iastate.edu} 
\affiliation{%
  \institution{Iowa State University}
  \city{Ames}
  \state{Iowa}
  \country{USA}
}

\author{Adarsh Krishnamurthy}
\email{adarsh@iastate.edu} 
\authornote{Corresponding authors.}
\affiliation{%
  \institution{Iowa State University}
  \city{Ames}
  \state{Iowa}
  \country{USA}
}

\renewcommand{\shortauthors}{Herron et al.}

\begin{abstract}
Recent advances in generative modeling, namely Diffusion models, have revolutionized generative modeling, enabling high-quality image generation tailored to user needs. This paper proposes a framework for the generative design of structural components. Specifically, we employ a Latent Diffusion model to generate potential designs of a component that can satisfy a set of problem-specific loading conditions. One of the distinct advantages our approach offers over other generative approaches, such as generative adversarial networks (GANs), is that it permits the editing of existing designs. We train our model using a dataset of geometries obtained from structural topology optimization utilizing the SIMP algorithm. Consequently, our framework generates inherently near-optimal designs. Our work presents quantitative results that support the structural performance of the generated designs and the variability in potential candidate designs. Furthermore, we provide evidence of the scalability of our framework by operating over voxel domains with resolutions varying from $32^3$ to $128^3$. Our framework can be used as a starting point for generating novel near-optimal designs similar to topology-optimized designs.
\end{abstract}


\begin{teaserfigure}
	\centering
    \includegraphics[trim=0 0 0 0,clip,width=0.9\linewidth]{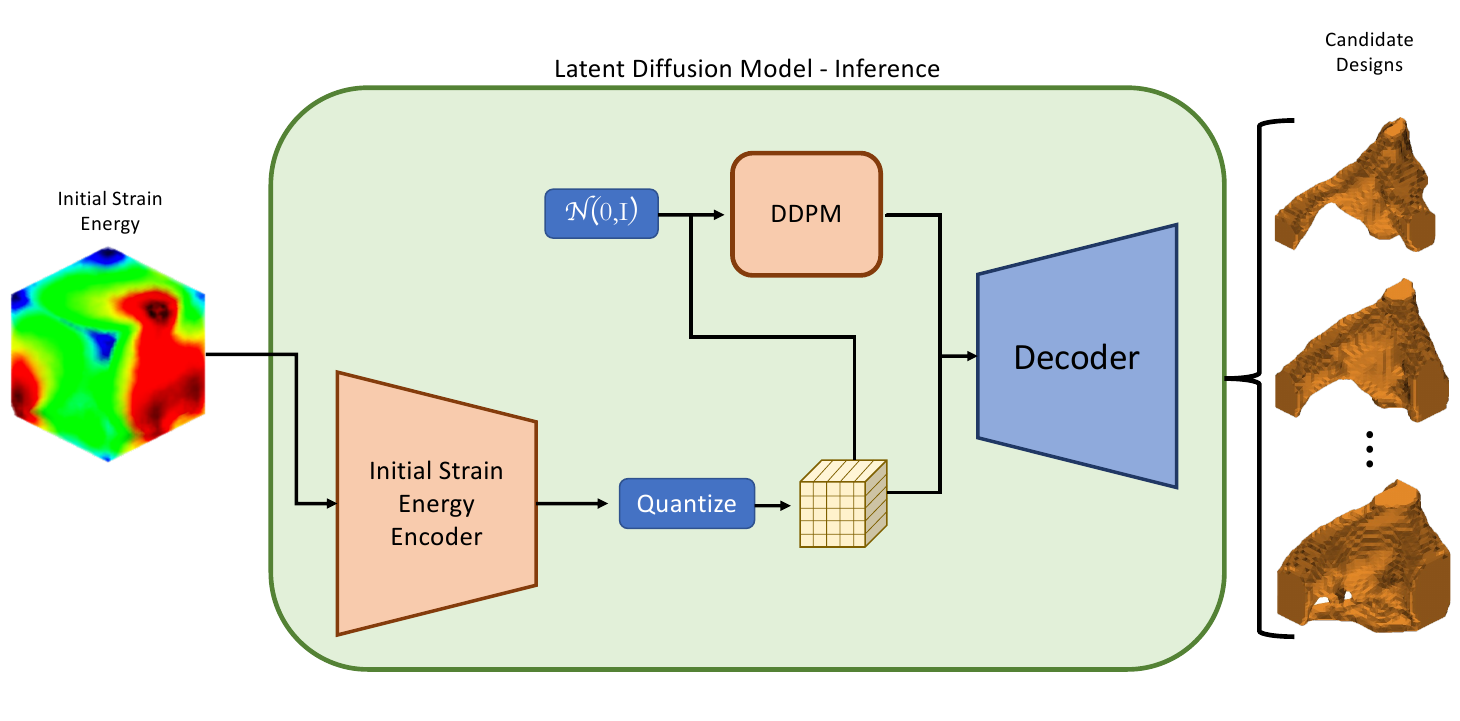}
    \caption{An overview of the proposed LDM framework for generating new designs (inference). Initial conditions in the form of a strain energy map are encoded to latent representations. The encoded initial conditions are used to condition the latent representations generated by the Diffusion Model. The encoded initial conditions and the generated latent representation are concatenated and decoded to render a potential candidate design. Different generated latent representations by the Diffusion Model will produce different candidate designs for a single set of initial conditions.}
	\label{fig:framework}
\end{teaserfigure}

\maketitle

\section{Introduction}

Computer-aided design (CAD) software allows engineers and designers to represent their designs digitally. Previously, engineers and designers relied on manual methods of representing and testing their designs, i.e., hand-drawn designs, physical prototypes, and manually calculating relevant computations. CAD software greatly simplifies the engineering process by representing designs digitally. Enabling designs to be easily shared among large teams, eliminating the need for elementary computations, and later, enabling the use of simulations, reducing the number of physical prototypes required. However, a specific bottleneck in the rapid design process is the initial design phase. Engineers and designers often create designs for a given system with only the general requirements in mind. Completing designs can also be incredibly time-consuming, particularly when designers can only create a few designs simultaneously. Multiple methods have been created to address this bottleneck; two of the most common methods are procedural modeling and topology optimization. Both methods have unique pros and cons, but they have the same goal: to improve the effectiveness of engineers and designers to speed up the design cycle. 

Recent advances in generative models, a subset of machine learning algorithms, provide the foundation for new methods in generative design. These new methods may build off of procedural modeling and topology optimization, allowing users to generate a set of potential candidate designs that may vary in overall design while maintaining structural performance. Multiple previous works have utilized generative models, Generative Adversarial Networks \citep{goodfellow2014generative, isola2018imagetoimage, kang2023gigagan} (GANs) in particular, for problems in design. Most of these works aim to develop surrogate models for current methods in topology optimization and do not address problem-specific exploratory design, i.e., generating a set of potential designs for a given set of initial conditions. This reduced capacity to generate sets of designs may be attributed to the generative algorithm used. GANs, in particular, may suffer from mode collapse, potentially outputting a single design for a set of initial conditions regardless of its inherent stochasticity. 

Diffusion Models (DMs) are a recent advance in generative modeling, which learns to denoise a data sample according to a predefined diffusion/noising process. DM's do not suffer from mode collapse at the expense of increased computational costs compared to GANs. In this work, the core of our framework is a Latent Diffusion Model (LDM). This architectural choice is expanded upon in \secref{Sec:Formulation}, but allows us to eliminate mode collapse while reducing the computational costs typically incurred by DMs. To train an LDM for problem-specific exploratory design, a distribution of designs parameterized in some manner is required. The optimally trained LDM can sample from this distribution of potential designs. We use the Solid Isotropic Material with Penalization (SIMP) topology optimization algorithm to generate this distribution of potential designs. The inherent stochasticity of the LDM enables it to generate a set of potential candidate designs conditioned on the initial conditions. A fortunate consequence of using a dataset generated with a Topology Optimization algorithm is that the designs generated by the LDM will be inherently near-optimal. We support this claim in \secref{Sec:Results} while elaborating on the common practices of evaluating generative models. 

This work addresses three major problems: reliably generating realistic near-optimal designs for a given set of initial conditions, generating \emph{multiple} candidate designs for a set of initial conditions, and reducing the computational costs of operating on 3D voxel grids. 

Our contributions reflect these major challenges and are as follows:
\begin{enumerate}
\item Develop a 3D generative design framework with Latent Diffusion Models, which enables the generation of structural components.
\item Train a series of Latent Diffusion Models on a dataset of approximately 66k $32^3$, 25k $64^3$, and 10k $128^3$ samples, encompassing diverse initial conditions and varying domain sizes.
\item Demonstrate the framework's ability to generate a set of diverse candidate structures for a given set of initial conditions with near-optimal structural performance.
\end{enumerate}

The paper is organized as follows. In \secref{Sec:Background}, we provide a background on generative design for structural components as well as generative modeling in terms of machine learning. Following this, we explain the mathematical formulations of Variational Autoencoders and Diffusion Models, the two major components of the proposed framework in \secref{Sec:Formulation}. These formulations of Variational Autoencoders and Diffusion Models are then put into context in \secref{Sec:Methodology}, where the overall framework is further elaborated. Finally, we include sections on data generation and analysis of our results in \secref{Sec:DataGen} and \secref{Sec:Results}, respectively.

\section{Background}
\label{Sec:Background}

One of the main pillars of the engineering design process is the conceptualization step, which includes design ideation and exploration. Typically, this process involves sketching, which is required to flesh out feasible designs, even after the invention of CAD. In line with all other technologies, considerable effort has been dedicated to automating this process. Topology optimization is the most deterministic method to automate this part or all of the design process. However, the objective of the topology optimization process is to find a specific optimal design of the part that would satisfy the different criteria. However, we would like to note that our problem formulation of generative design is distinct from topology optimization. The proposed framework is focused on generating \emph{multiple} different candidate designs for a given set of initial conditions with the added benefit of the designs being inherently near optimal. 

Current methods of generative shape design for structural components are dominated by computationally expensive optimization algorithms~\citep{orme2017designing,liu2016survey,app13010479}, such as Solid Isotropic Material with Penalization (SIMP) method~\citep{bendsoe1988generating, bendsoe1989optimal}, Level Sets~\citep{wang2003level,van2010level}, and Genetic algorithms~\cite{Hajela1993, WANG20053749}. Typically, the objective function for which these methods optimize is the design's total strain energy or compliance accompanied by user-defined constraints. These methods are computationally expensive due to the iterative requirements and finite element (FE) evaluations required to compute the objective function. Out of these methods, genetic algorithms are the only ones to provide multiple different candidate designs for a set of initial conditions, while the other methods are deterministic. Regardless, generating new potential designs on the fly is infeasible due to the FE solution calls. Reducing the computational costs of these methods would provide substantial value to design engineers from any domain. 

Generative modeling is a statistical term for modeling a joint probability distribution, $P(X, Y)$, with the observable variable being $X$ and the target variable being $Y$. This can be extremely difficult to model, particularly as the sample size defining the distribution of $Y$ grows, quickly making it intractable. In this work, \emph{generative modeling} refers to methods that use neural networks to approximate the joint probability distribution, $P(X, Y)$.

Three of these generative modeling methods are GANs, Variational Autoencoders (VAE), and Diffusion Models (DM), more formally, Denoising Diffusion Probabilistic Models (DDPMs). From a high level, these methods fall into two categories of generative modeling: explicit and implicit. VAEs and DMs explicitly define the joint probability distribution, optimizing the neural networks to maximize the variational lower bound. GANs implicitly model the joint distribution by optimizing a secondary neural network to learn the difference between samples in the target distribution and those out of the target distribution. Irrespective of the objective function formulation, each method learns a mapping from a known prior distribution $P(X)$ to the unknown, high dimensional, and complex target distribution $P(Y)$. Typically, a Normal Gaussian distribution is chosen as the prior distribution, and the target distribution is problem-specific. Each of these methods has its unique pros and cons. GANs have shown remarkable results and are computationally economical but are prone to mode collapse and instabilities during training. VAEs and DMs do not suffer from mode collapse or instabilities during training. VAEs often fall to the mean of the dataset and generate samples dominated by lower-frequency features, resulting in blurry samples (in the case of image generation). DMs can model extremely high-frequency features at the expense of increased computational costs. This trade-off comes from the iterative refinement nature of DMs. DMs are trained to remove isotropic Gaussian noise from a noisy data sample iteratively. During inference, the DM starts from pure isotropic Gaussian noise, iteratively removing noise until it reaches an approximate sample from the target distribution. A consequence of this iterative refinement structure is the DM can denoise from any arbitrary point in the diffusion/noising process~\citep{saharia2022palette}. In this work, denoising a partially noised sample amounts to editing an existing design, a desirable functionality for designers. 

Current state-of-the-art (SOTA) generative models for image generation unite VAEs and DMs in LDMs. LDMs are trained with a two-stage process. First, a VAE is trained on the target dataset, learning a lower-dimensional representation of the target distribution in its latent space. Second, a DM is trained to denoise latent samples from the pretrained VAE. So, the ground truth samples used for the DM's training are encoded latent representations from the pretrained VAE. This LDM formulation reduces the computational costs incurred from running the DM while improving the performance of the VAE. 

The most related works, in terms of machine learning for structural component design, take the topology optimization (TO) approach rather than generative design. Several initial approaches utilize supervised learning methods, amounting to fast surrogate models for the computationally demanding TO algorithms~\citep{Woldseth_2022,sosnovik2019neural,banga20183d,yu2019deep,zhang2019deep,chandrasekhar2020tounn, qiyinLin,kollmann2dmetamaterials, rawat2019novel,Yu_2018,HAMDIA201921,shuheidoi, lagaros, RADE2021104483,oh_2019,baotongLi, diabWAbueidda,saskaiIPMmotor,zhang2020deep, oh2018GANTO,zhou2020,guo2018, leeSeunghye,de2019topology,bujny2018,takahashi2019convolutional, chaoQian2020,jang2020generative,rodriguez2021improved,poma2020optimization, CHI2021112739}. These supervised approaches produced impressive results with dramatic speed-ups and minimal loss in performance to their respective TO algorithms. A consequence of using supervised methods is the lack of stochasticity, strictly producing one-to-one mappings for sets of initial conditions. Subsequent methods utilize generative models, VAEs, GANs, and even a DM, all with impressive results~\citep{Nie2020TopoGAN, bernhard2021topogan, mazé2022diffusion}. These methods heavily focus on the performance of the generated component, optimizing the networks to match their TO counterpart or surpass it with added performance terms in the objective function. Most previous works outlined above operated in 2D space with relatively small and constrained datasets, e.g., traditional cantilever TO problem. 


\section{Mathematical Formulation}\label{Sec:Formulation}
The core of our framework is a Latent Diffusion Model. LDMs are composed of two main components, an external VAE, and an internal DM, which operate in the latent space of the VAE, hence \emph{Latent} Diffusion Model. The following section covers the mathematical preliminaries of VAEs followed by DMs.

\subsection{Variational Autoencoders}
\label{SubSec:VAE}

Unlike traditional Autoencoders, which map an input $\mathbf{x}$ to a latent representation $\mathbf{z}$, VAEs map an input $\mathbf{x}$ to a probability distribution. The VAE's encoder defines this probability distribution by predicting its mean and variance. The latent variable $\mathbf{z}$ is extracted from this distribution by randomly sampling a standard normal Gaussian, multiplying it with the predicted variance, and shifting that product by the predicted mean. More formally:
\begin{equation}
    \mathbf{z} = \mathbf{\mu} + \mathbf{\sigma} \odot \mathbf{\epsilon}
\end{equation}
where $\mathbf{\epsilon} \sim \mathcal{N}(0,\mathbf{I})$ and $\odot$ is element-wise multiplication. The VAE's decoder then predicts $\hat{\mathbf{x}}$, a reconstruction of the original sample $\mathbf{x}$, given the random sample $\mathbf{z}$ from the distribution defined by the encoder. In summary, the VAE is composed of an encoder $p_{\theta}(\mathbf{z}|\mathbf{x})$, the random latent sample $\mathbf{z}$ which comes from a prior distribution $p_{\theta}(\mathbf{z})$, and a decoder $p_{\theta}(\mathbf{x}|\mathbf{z})$. The encoder and decoder are both neural networks parameterized by $\theta$. By formulating the VAE such that the latent variable $\mathbf{z}$ is sampled from some prior distribution, during inference, the decoder can then randomly sample from the target distribution by decoding a latent variable $\mathbf{z}$, which is drawn from a standard normal distribution. This ability qualifies the VAE as a generative modeling algorithm. 

During training, the VAE is optimized with two different loss terms; the first is a simple reconstruction loss, similar to a traditional autoencoder, and the second term is for the latent variable $\mathbf{z}$. This second term is used to regularize the latent variable sampled from the distribution defined by the encoder towards the prior distribution, a standard normal Gaussian. This is accomplished by minimizing the KL divergence between the prior distribution, $p_{\theta}(\mathbf{z})$, and the distribution defined by the encoder, $p_{\theta}(\mathbf{z}|\mathbf{x})$. Thus, our VAE loss function is:
\begin{equation}
    \mathcal{L}(\theta; \mathbf{x}) = \mathbb{E}_{\mathbf{z} \sim p_{\theta}(\mathbf{z}|\mathbf{x})}\Big[ \|\mathbf{x} - p_{\theta}(\mathbf{x}|\mathbf{z}) \|\Big] + KL\Big(p_{\theta}(\mathbf{z}|\mathbf{x}) || p(\mathbf{z})\Big)
\end{equation}

\subsection{Vector-Quantized Variational Autoencoders}
\label{SubSec:VQVAE}

A Vector Quantized VAE (VQ-VAE) \citep{oord2018neural} is a VAE that uses a discrete codebook of embeddings to quantize the continuous latent space learned by a standard VAE. Opposed to a normal Gaussian distribution defining the VAE's latent prior distribution, $p(\mathbf{z})$, the discrete codebook of embeddings defines the VQ-VAE's prior distribution. The discrete codebook contains $K$ embeddings, each of dimension $\mathbb{R}^D$, that act as the nearest discrete approximation to the continuous latent space. During the encoding step, the input $\mathbf{x}$ is first mapped to the continuous latent space by the encoder function $q(\mathbf{z}|\mathbf{x})$. Then, the continuous latent variable is quantized by finding the nearest embedding, by Euclidean distance, in the codebook. The decoder function $p(\mathbf{x}|\mathbf{z}_q)$ decodes the embeddings, $\mathbf{z}_q$,  to reconstruct the input $\mathbf{x}$. The decoder will only ever see embeddings from the discrete codebook, but the combination of discrete codes will vary across different samples. Overall, the VQ-VAE can be seen as a trade-off between the expressive power of a continuous latent space and the efficiency of a discrete codebook.

The VQ-VAE is trained with the same loss function as a VAE, except instead of minimizing the KL-Divergence between the predicted latent variable $\mathbf{z}$, the VQ-VAE is optimized to minimize the Euclidean distance between the predicted latent codes and its nearest, in Euclidean distance, discrete code.

\subsection{Denoising Diffusion Probabilistic Models (DDPM)}
Diffusion models comprise two distinct parts: the forward and backward diffusion processes. The forward diffusion process is simply the iterative addition of Gaussian noise to a sample from some target distribution. This forward process may be carried out an arbitrary number of times, creating a set of progressively noisier and noisier samples of the original data sample. The following Markov chain defines this process:
\begin{equation}
    q(\mathbf{x}_t | \mathbf{x}_{t-1}) = \mathcal{N}(\mathbf{x}_t; \sqrt{1 - \beta_t}\mathbf{x}_{t-1},\beta_t\mathbf{I}),
\end{equation}
where $ \mathbf{x}_0 $ is the sample from some target distribution $ q(\mathbf{x}) $, and a variance schedule defined as $ \{ \beta_t \in (0, 1) \}^T _{t=1} $. The reverse diffusion process would then iteratively remove the noise added during the forward diffusion process, i.e., $ q(\mathbf{x}_{t-1}|\mathbf{x}_t) $. 

In practice, it isn't feasible to sample $ q(\mathbf{x}_{t-1}|\mathbf{x}_t) $ since we do not know the entire distribution. Therefore, we approximate the conditional probabilities via a neural network, $ G_\theta(\mathbf{x}_{t-1}|\mathbf{x}_t) $, where the parameters of $ G$, and $ \theta $, are updated via gradient-based optimization. Training a diffusion model then consists of optimizing a neural network, $ G_\theta(\mathbf{x_t}, t) $, to reconstruct the random normal noise used in the forward diffusion process to transform the original sample into a noisy sample ($\mathbf{x_t}$) at an arbitrary timestep in the diffusion process. This objective function is defined as:
\begin{equation}
    \|z - G_{\theta}(\mathbf{x}_t, t)\|^2 = \| z - G_{\theta}(\sqrt{\Bar{\alpha}_t}\mathbf{x}_0 + \sqrt{(1 - \Bar{\alpha}_t)}z, t) \|^2
\end{equation}
where $ \alpha_t = 1 - \beta_t $, $ \Bar{\alpha}_t = \Pi^t _{s=1}\alpha_s $ and $ z \sim \mathcal{N}(0,I) $. 

In summary, the neural network is given a noisy data sample at some arbitrary timestep in the diffusion process and predicts the randomly sampled normal Gaussian noise used for the forward diffusion process. Since the KL divergence between two standard normal Gaussian distributions is the $\mathcal{L}_2$ loss, we can optimize our neural network using a simple Mean Squared Error reconstruction loss.

\section{Methodology}
\label{Sec:Methodology}

In this work, we develop a framework that maps the initial conditions to a diverse range of potential candidate designs, contingent upon the specified constraints. This problem is particularly well-suited for a generative modeling algorithm, given that multiple near-optimal or satisfactory solutions may exist. We choose to employ a Latent Diffusion Model owing to its advantages in training stability, distribution coverage (emphasizing mode coverage), and computational efficiency. As previously discussed, the Latent Diffusion Model comprises two distinct components: an external Variational Autoencoder (VAE) and an internal Diffusion Model (DM). In the following subsections, we provide a detailed account of the design choices made for each component and an overview of the framework's capabilities, thereby demonstrating its suitability for application in structural component design.

The final implementation of our framework may be seen in \figref{fig:framework}. At inference, the input to the framework is the problem-specific initial conditions in the form of a 3D voxel grid. The output is a binary 3D voxel grid, where 1s denote material presence, and 0s denote no material presence. The multiple candidate designs emphasize the ability of the framework to generate different structures for the same initial conditions given different initial random vectors. \figref{fig:training_vae} outlines the training process of the VAE where the input is a tuple of the initial conditions and SIMP-Optimized structure, both of which are 3D voxel grids of the same size. In this figure, it can be seen the VAE is trained to reconstruct the SIMP-Optimized structure used as input to the \emph{Design Encoder}. Lastly, a visual of the LDM training loop may be found in \figref{fig:training_LDM} where the Multi-Headed VAE's Decoder is absent, and the Diffusion Model is optimized to invert the noising process used on the \emph{Design Encoder's} latent representation. 

The dataset employed for training and validation comprises pairs of initial conditions and corresponding ground truth: designs optimized using the Solid Isotropic Material with Penalization (SIMP) method. The initial conditions are represented as a single voxel grid, where each voxel contains continuous normalized strain energy values. A comprehensive explanation of the dataset generation process can be found in \secref{Sec:DataGen}.

\subsection{Multi-Headed Variational Autoencoder}
\label{SubSec:MHVAE}

\begin{figure}[t!]
	\centering
    \includegraphics[trim=0 0 0 0,clip,width=0.98\linewidth]{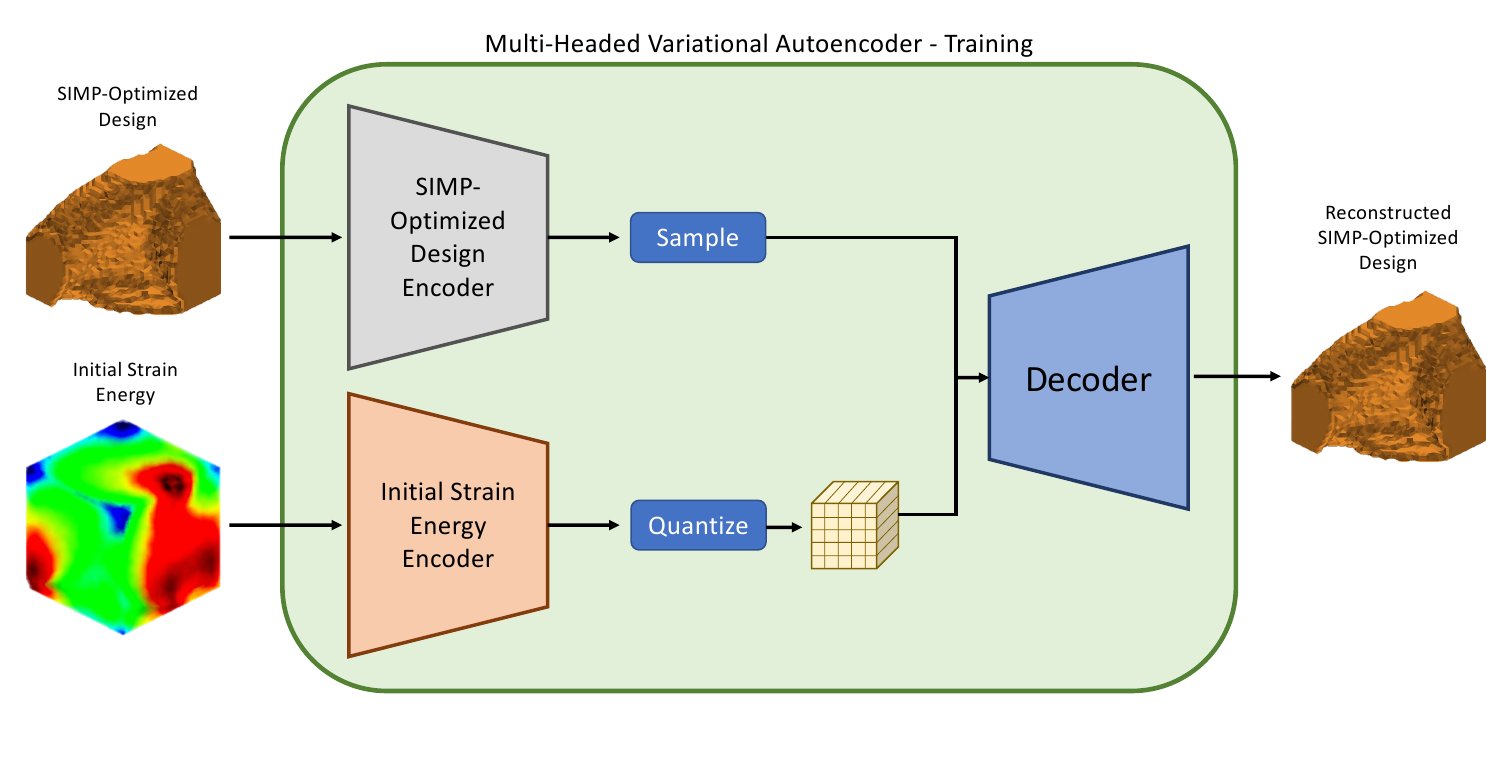}
    \caption{An overview of training the external Multi-Headed VAE.}
	\label{fig:training_vae}
\end{figure}

The external VAE of our framework is composed of two different encoder networks and a single decoder network. We further refer to the external VAE as a multi-headed VAE. One encoder (head) compresses the Initial Strain Energy into its respective latent representation, while the other compresses the SIMP-Optimized density into its respective latent representation. Each latent representation is a tensor of size $[4,8,8,8]$. The decoder upsamples the concatenated latent representations (total size of $[8,8,8,8]$) back to the voxel space. When given the Initial Strain Energy and SIMP-Optimized design, the entire multi-headed VAE is optimized to reconstruct the SIMP-Optimized design, essentially a traditional VAE conditioned on the Initial Strain Energy. This training process is shown in \figref{fig:training_vae}.

In our multi-headed VAE formulation, we implement two different regularization loss terms, one for each encoder's respective latent representation. The Initial Strain Energy encoder utilizes a Vector-Quantized bottleneck (VQ-VAE). In contrast, the Optimized Design encoder uses the original KL-Divergence formulation for its latent regularization term. Preliminary experimental results found the VQ-KL formulation of the multi-headed VAE, VQ bottleneck for initial conditions and KL bottleneck for Optimized Designs, produced better results in terms of connected components, e.g., VQ bottlenecks for each encoder was prone to generating designs with floating artifacts. Further exploration of these architectural choices is left for future works.

\subsection{Latent Diffusion Model}
\label{SubSec:LDM}

During inference, our framework may only have access to the Initial Strain Energy when generating a new candidate design. A generative model is then required to generate a latent representation that appears to come from the distribution defined by the Optimized-Density encoder. This requirement motivates using a DM in the latent space of the multi-headed VAE. The generated latent representation is concatenated with the latent representation from the Initial Strain Energy encoder and decoded back to the voxel space, rendering a potential candidate design.

\begin{figure}[t!]
	\centering
    \includegraphics[trim=0 0 0 0,clip,width=0.98\linewidth]{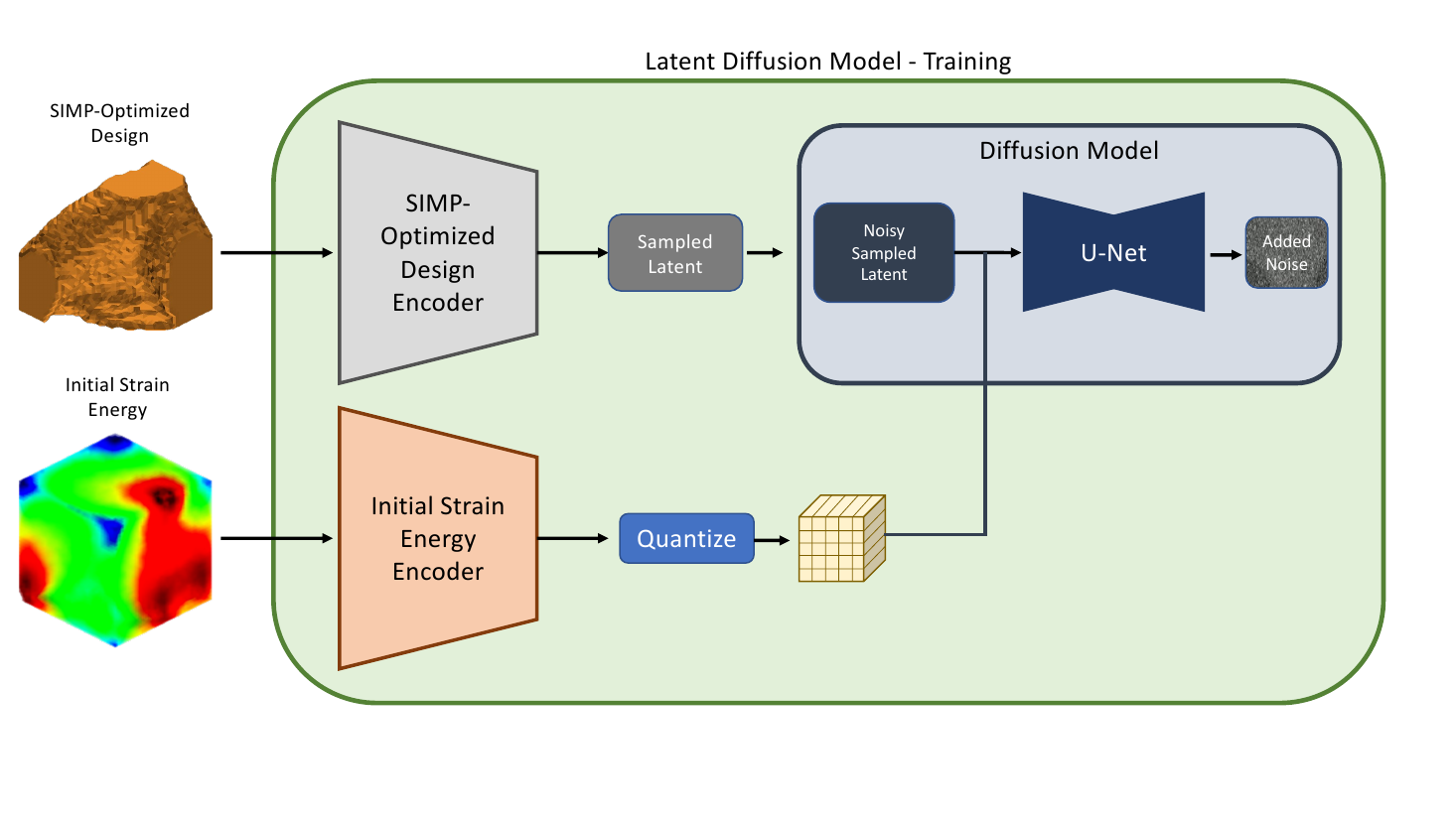}
    \caption{An overview of training the Latent Diffusion Model.}
	\label{fig:training_LDM}
\end{figure}

This DM is trained and sampled like a traditional DDPM, learning to reconstruct the pure white noise used to compute the input noisy sample. Since this is a DM operating in the latent space of the VAE, the ground-truth samples are the compressed latent representations by the \textbf{pretrained} multi-headed VAE. The DM is trained to reconstruct the pure white noise used to compute the noisy latent representations sampled from the Optimized Design encoder's posterior distribution. Additionally, the DM is conditioned on the latent representation from the Initial Strain Energy encoder. We condition this DM by concatenating the noisy Optimized Design latent with the unperturbed Initial Strain Energy latent as input to the DM. The DM is then learning a mapping of, $\mathbb{R}^{[8,8,8,8]} \rightarrow \mathbb{R}^{[4,8,8,8]}$, where the input is a clean conditioning sample concatenated with the noisy sample to predict the pure noise used to get the noisy sample. Recently, conditional DMs have used classifier-free guidance~\citep{ho2022classifierfree}. Due to the specific application of our proposed framework, we do not implement any traditional form of guidance or conditioning, i.e., classifier or classifier-free. We found that always providing the DM with latent representations of the initial conditions sufficiently conditioned sample generation. The training process of the DM may be seen in \figref{fig:training_LDM}. 

\begin{figure*}[t!]
	\centering
    \includegraphics[trim=0 0 0 0,clip,width=0.99\linewidth]{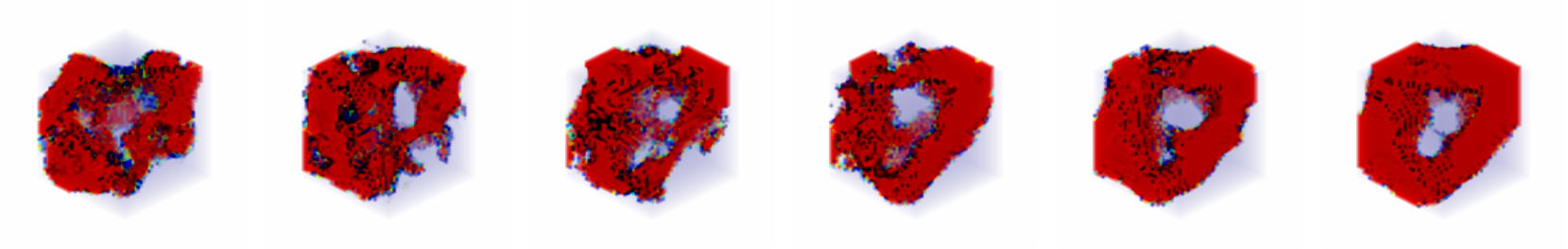}
    \caption{Decoded posteriors from the DDPM denoising trajectory during inference. Decoding the initial random Gaussian sample (far left design) highlights the improvement our framework provides over a traditional VAE.}
	\label{fig:trajectory}
\end{figure*}

\subsection{Design Generation and Translation}
Our framework offers two primary capabilities: design generation and design translation. Design generation is the direct inference process described earlier, wherein an initial strain energy is an input, an optimized design latent is generated, and a candidate design is subsequently decoded from this latent representation.

Design translation serves as a secondary capability of our framework, necessitating no additional training, and addresses the question: Given an initial design, can we generate several inherently similar yet distinct designs? In this process, the initial strain energy and design are mapped to their respective latent representations using their corresponding encoders. The Diffusion Model is subsequently employed to iteratively denoise a partially noised latent representation of the initial design. It is important to note that the underlying structure is preserved by only partially noising the latent representation, while higher frequency features are removed. The variability in the edited latent representation is approximately proportional to the degree of initial noise introduced; a noisier latent representation results in a more distinct translated design. For a visual example of this process, please refer to \figref{fig:exampleDesignTranslation}.

\section{Data Generation}
\label{Sec:DataGen}

We use the ANSYS Mechanical APDL v19.2 software to generate the dataset, which uses the SIMP method for performing structural topology optimization. We use a mesh of a cube with a length of 1 meter as an initial geometry. The mesh of the cube consists of 31093 nodes and 154,677 elements with 8 degrees of freedom. To guarantee that the dataset contains diverse shapes originating from the cube, we use a set of different loading and boundary conditions available in ANSYS, like Nodal Force, Surface Force, Remote Force, Pressure, Moment, and Displacement. To create a single topology optimization geometry, we use a random selection process to choose three nodes that are not collinear and designate them as fixed points by assigning them zero displacements. The type of load to be applied is also randomly selected from the available loads within the ANSYS software, and the location on the mesh where the load is to be applied is chosen randomly as well. The specific value of the load is then determined by sampling from the predefined range assigned to that particular load type. Additionally, to generate topologies of varying shapes, we utilize a range of target volume fractions with a minimum value of 10\% and a maximum of up to 50\%, with an increment of 5\%. We generated 66,000 (66K) samples by utilizing the randomization above scheme. 

We save the initial strain energy and optimal topology in a mesh format throughout the data generation phase. However, to train the 3D CNNs, it is necessary to transform the mesh data into a voxel representation. We adopt the voxelization process outlined in~\citep{RADE2021104483} to convert all samples to voxel-based representations. We choose three different resolutions for voxelization: $32^3$, $64^3$, and $128^3$. The dataset is divided into training and testing sets, consisting of 80\% and 20\% of all samples, respectively. The voxelization process, which takes around 15 minutes to complete for a single geometry, was sped up significantly by using GNU parallel to perform the voxelization of all samples on the cluster.

\section{Results}
\label{Sec:Results}

In the following section, we present several different modes of analysis. These analyses are conducted on a small subset of the test set, consisting of 100 sample pairs of Initial Strain Energy and SIMP-Optimized designs. For each of the 100 different samples, we generated 20 unique designs corresponding to their respective Initial Strain Energy. This evaluation set includes 2,000 LDM-Generated designs and 100 SIMP-Optimized designs. 

Additionally, we would like to highlight common practice metrics for generative modeling algorithms. Due to the nature of generative modeling (approximating intractable high-dimensional probability distributions), there isn't a clean or easily computable performance metric. It is then common practice to evaluate the performance of generative models with the \emph{Frechet Inception Distance}~\citep{NIPS2017_8a1d6947} (FID). The FID score is obtained by comparing the mean and standard deviation of two datasets, generated and ground truth images. Each sample in each dataset is the output of the deepest layer in a pretrained Inception-v3~\citep{szegedy2015rethinking} model. Inception-v3 model is pretrained as an ImageNet classifier, meaning any design, ground truth, or generated model would be Out of Distribution (OOD), as well as 3D, not a 2D image, rendering the FID score meaningless for evaluating our work.

To quantitatively evaluate our model we train an evaluation network that predicts the total amount of strain energy for a given design, specific details are discussed in \secref{Results:Strain Energy Analysis}. The evaluation network serves as an application-specific 'Inception-v3' model, providing a scalar metric to compare the generated designs with ground truth designs. 

\begin{figure}[t!]
	\centering
    \includegraphics[trim=0 0 0 0,clip,width=0.9\linewidth]{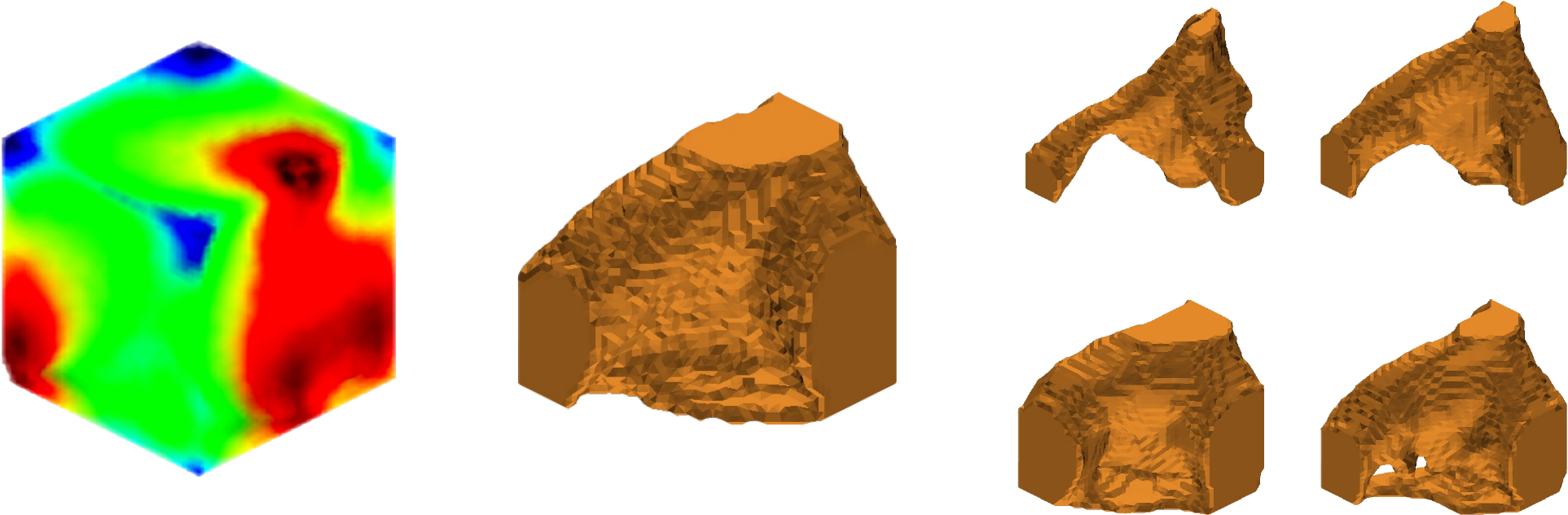}
    \caption{Multiple candidate designs for a given Initial Strain Energy Map. From left to right, Initial Strain Energy, SIMP-Optimized Design, and four different generated designs by the proposed framework.}
	\label{fig:exampleCandidateDesigns}
\end{figure}

\subsection{Design Generation}
\label{SubSec:DesignGen}

\begin{figure}[t!]
    \centering
    \includegraphics[width=\linewidth]{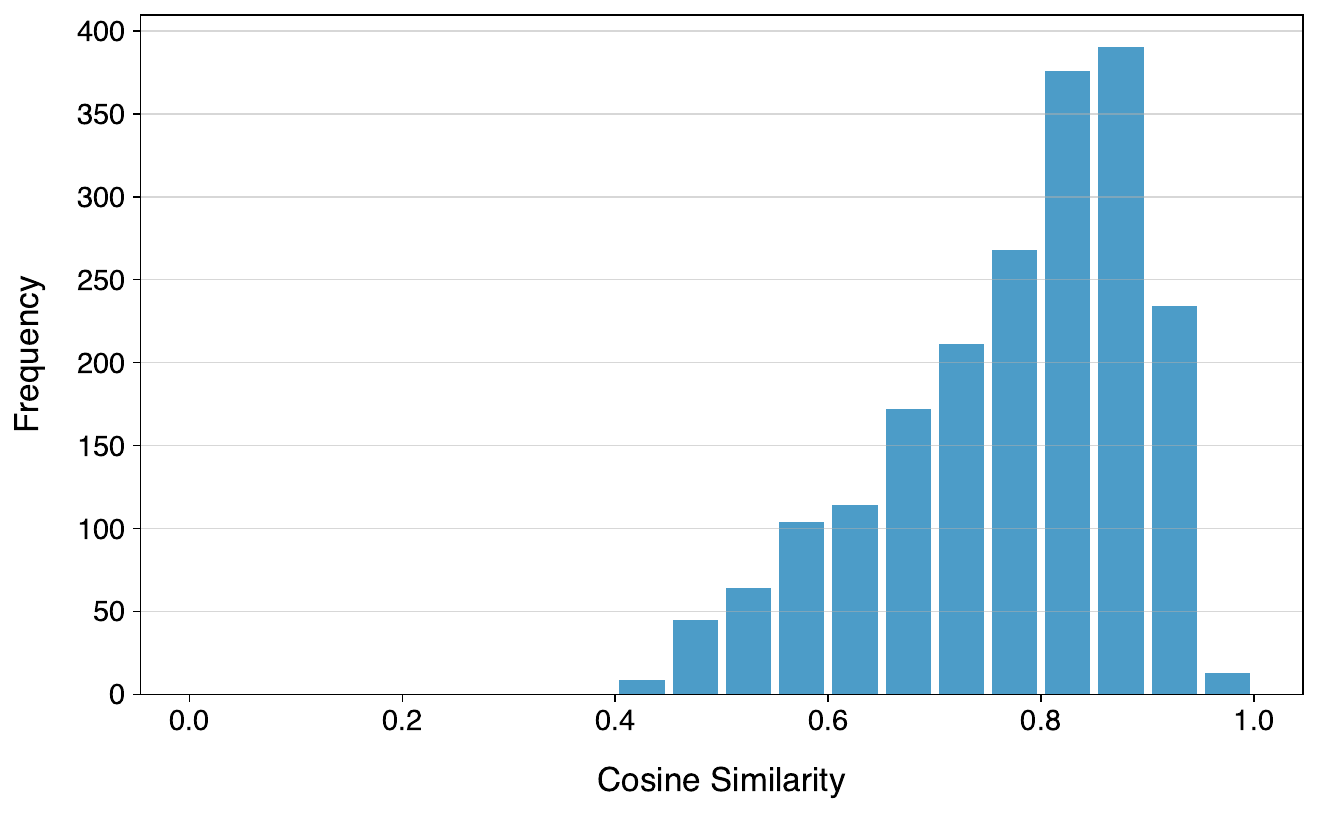}
    \caption{The distribution of cosine similarity between the SIMP-Optimized design and LDM-Generated designs for a given Initial Strain Energy.}
    \label{fig:total_cossim}
\end{figure}

\begin{figure*}[t!]
    \centering
    \begin{subfigure}{0.46\linewidth}
        \includegraphics[width=0.95\linewidth,trim={2.5in, 1.5in, 2.5in, 1.5in},clip]{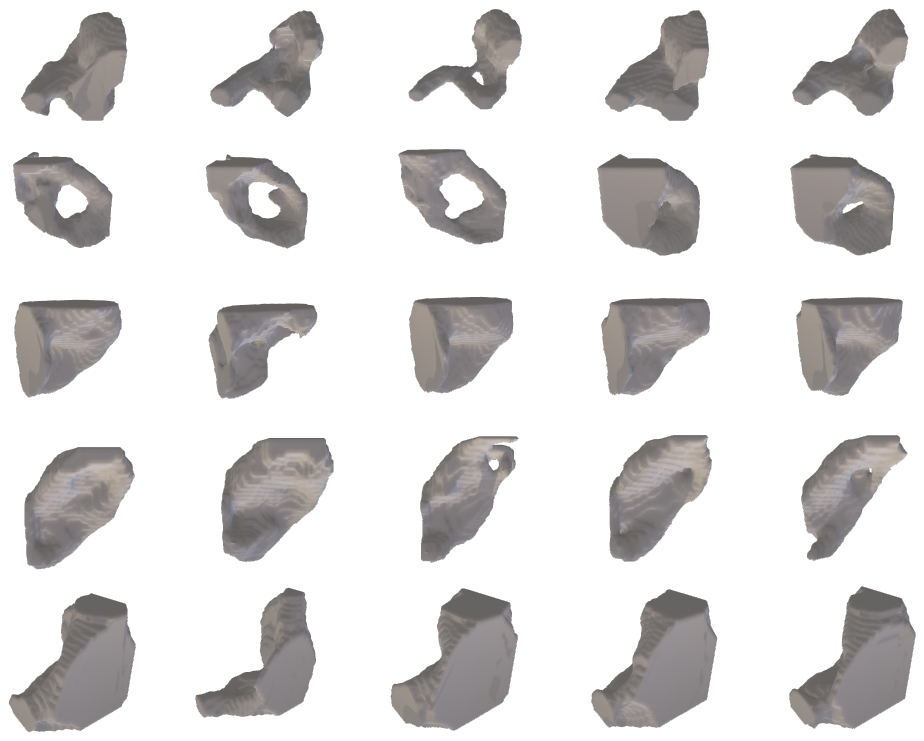}
        \caption{$32^3$ resolution}
        \label{fig:designs_32cube}
    \end{subfigure}
    \hspace{0.06\linewidth}
    \begin{subfigure}{0.46\linewidth}
        \includegraphics[width=0.95\linewidth,trim={2.5in, 1.5in, 2.5in, 1.5in},clip]{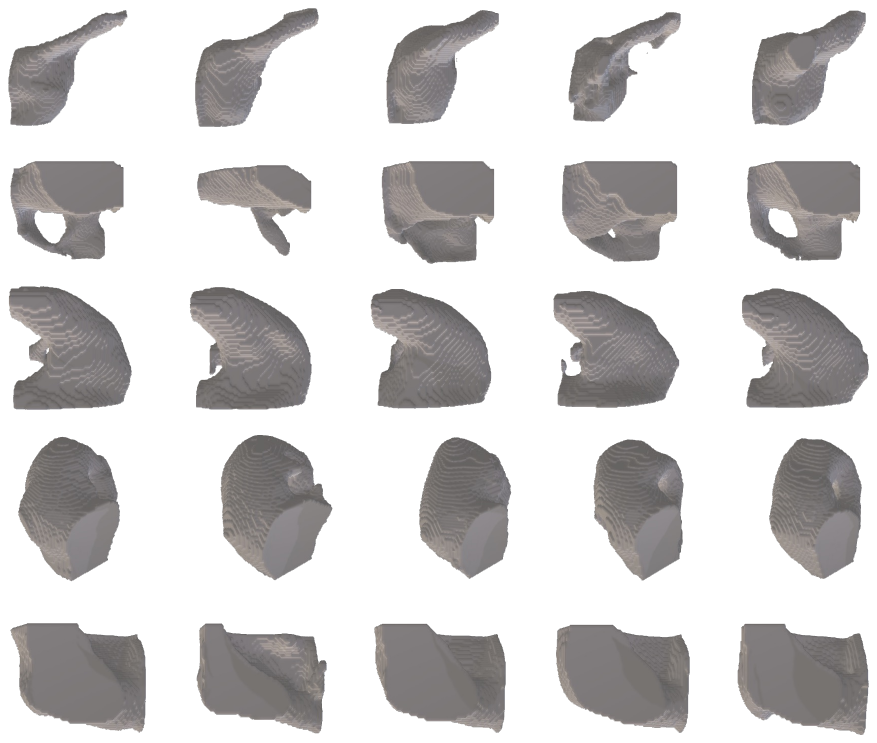}
        \caption{$64^3$ resolution}
        \label{fig:designs_64cube}
    \end{subfigure}
    \caption{Example generated designs. Each row corresponds to a single Initial Strain Energy sample, and each column is a different generated design.}
\end{figure*}

As a reminder, the design generation task is the fundamental capability of our framework, which maps a given Initial Strain Energy to a potential candidate design. Due to the stochasticity of the DM, every different initial random sample the DM starts with will render a unique design. We evaluate the performance of our framework qualitatively alongside quantitative evaluations to support the efficacy and capabilities of the framework. As the main contribution of this paper is generative design, we qualitatively assess the model's performance by human perception. Given an Initial Strain Energy, does the framework provide multiple unique designs that seem feasible and potentially aesthetically appealing? From \figref{fig:exampleCandidateDesigns} and \figref{fig:designs_32cube}, it is apparent that the framework can generate multiple different candidate designs, all with substantial variability. To evaluate this idea of variability with a quantitative metric, we compute the Cosine Similarity between the 20 LDM-Generated designs and their corresponding SIMP-Optimized design for all 100 test samples. This histogram may be seen in \figref{fig:total_cossim}, where we can conclude the LDM consistently generates unique designs and does not simply reconstruct the corresponding SIMP-Optimized design.

\begin{figure}[t!]
    \centering
    \includegraphics[width=\linewidth]{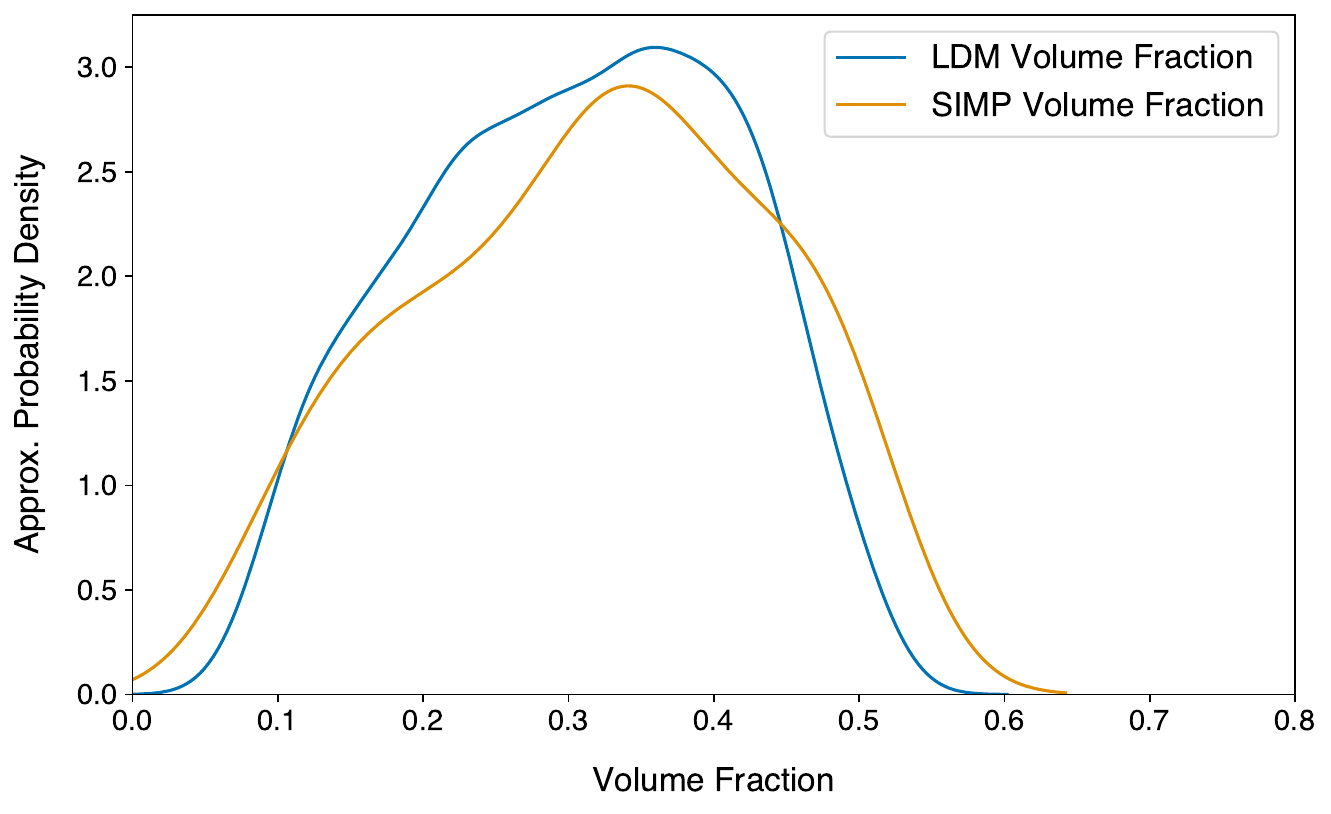}
    \caption{The distributions of volume fractions for SIMP-Optimized designs and LDM-Generated designs.}
    \label{fig:vf}
\end{figure}

\begin{figure}[t!]
    \centering
    \includegraphics[width=\linewidth]{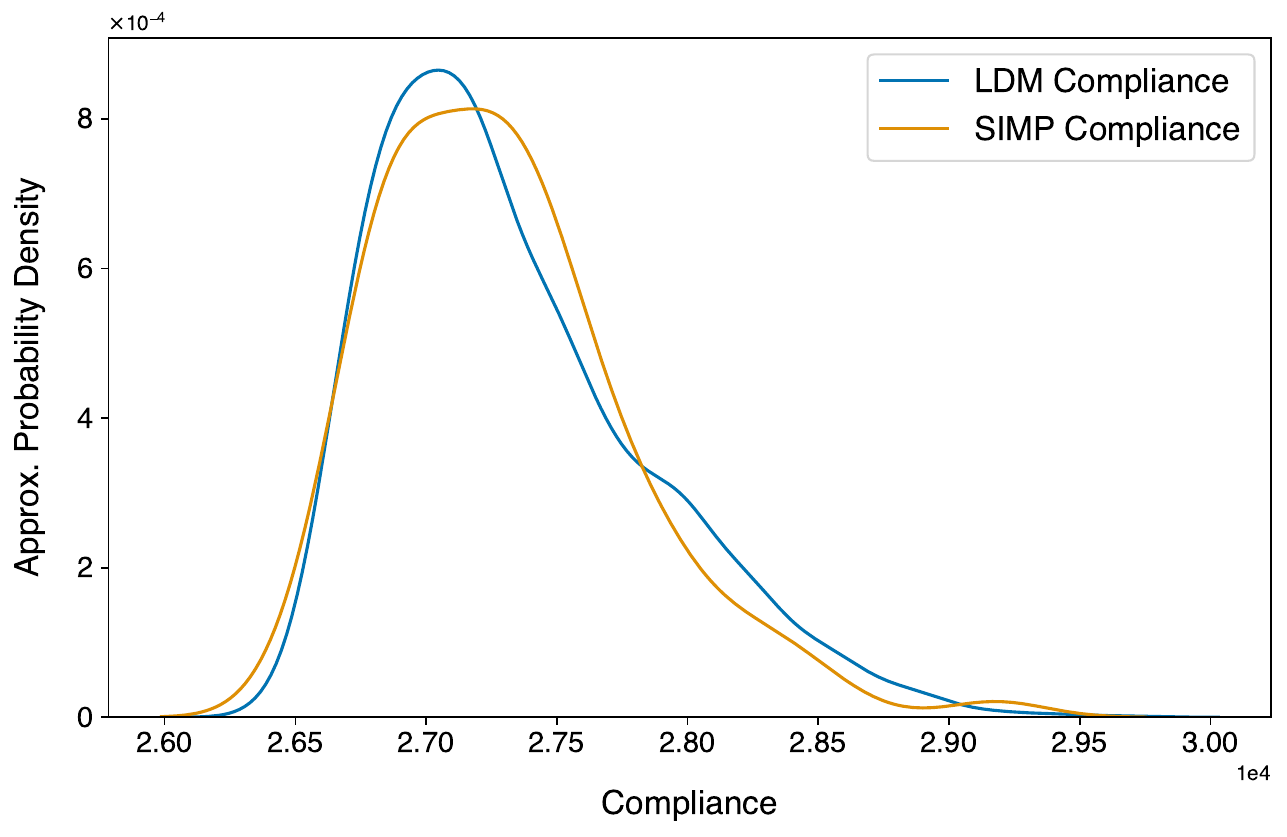}
    \caption{The distributions of predicted strain energy for SIMP-Optimized designs and LDM-Generated designs.}
    \label{fig:total_kde}
\end{figure}

\begin{figure*}[t!]
	\centering
    \includegraphics[width=0.8\linewidth,trim=1.4in 1.5in 0.8in 1.5in,clip]{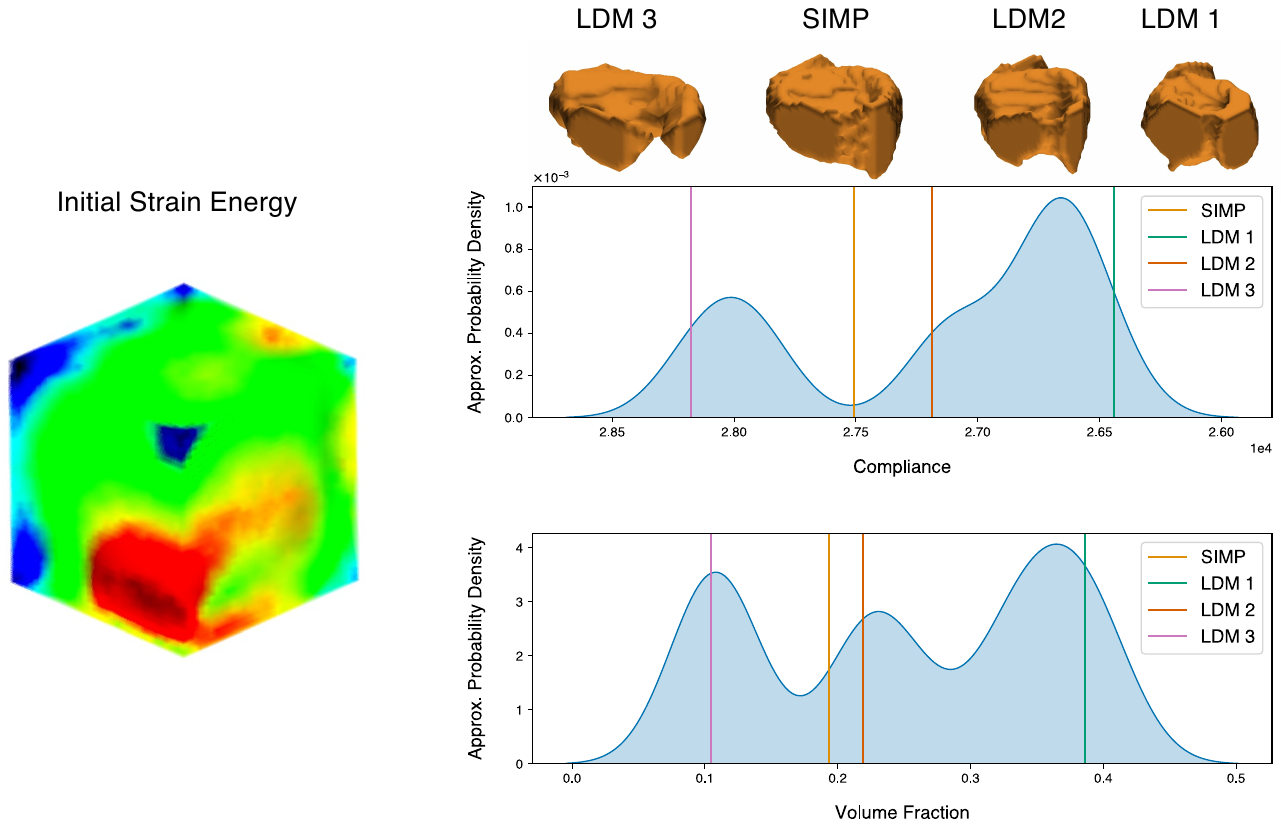}
    \caption{Single Instance KDE plot comparing potential designs and the SIMP optimized design relative to the distribution of predicted compliance values (top) and volume fractions (bottom). The Initial Strain Energy for this sample is shown on the left. The designs, generated and SIMP-Optimized, are placed according to their predicted compliance values.}
	\label{fig:KDE_SI_Figure}
\end{figure*}

\begin{figure*}[t!]
    \centering
    \includegraphics[width=0.95\linewidth]{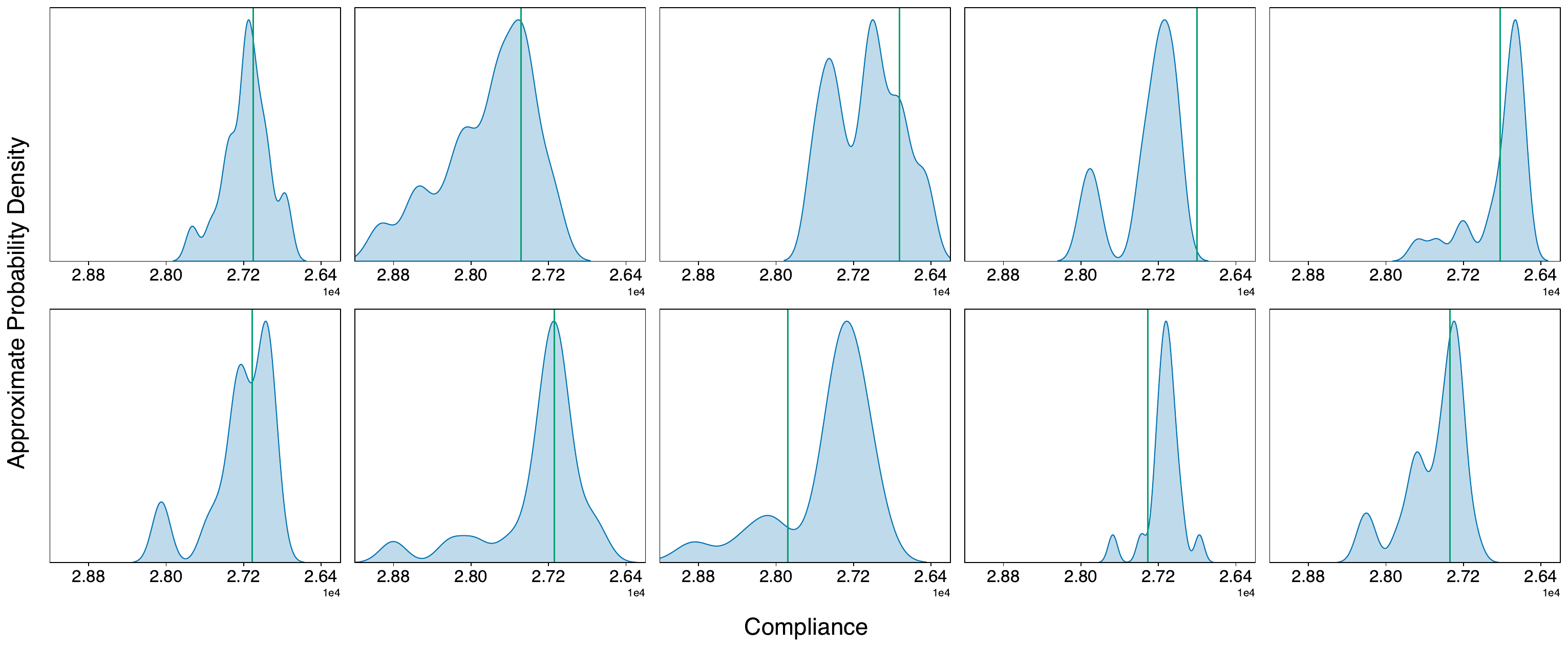}
    \caption{Each subfigure is the predicted total strain energy distribution for 20 generated designs for a given Initial Strain Energy. The red bar is the predicted total strain energy of the corresponding SIMP-Optimized design.}
    \label{fig:kde_subplots}
\end{figure*}

In \figref{fig:trajectory}, we decode several latent representations during the reverse diffusion process undertaken by the DM during inference. If the DM were operating in voxel space, this graphic would show the mapping from pure white noise to a coherent design. Since the DM is operating in the latent space of the multi-headed VAE, we need to decode the latent representations along the reverse diffusion trajectory to visualize this process. In doing so, we can see snapshots from the iterative refinement trajectory taken by the DM. \figref{fig:trajectory} also highlights the improvement a LDM provides over a traditional VAE. In a traditional VAE, a randomly sampled Gaussian is decoded; this decoded design is the far left figure, which contains a vague overall structure lacking accurate high-frequency features. In summary, this figure directly compares the results for a traditional VAE and our framework, the far left vs. the far right design.

\subsection{Strain Energy Analysis}
\label{Results:Strain Energy Analysis}

The added benefit of training the VAE on SIMP-Optimized designs is that the generated designs by our framework are inherently near-optimal. To support this claim, we trained a surrogate neural network, the aforementioned evaluation network, to predict the strain energy of a design given the Initial Strain Energy. To train a surrogate network, we need the dataset of strain energies and topologies at intermediate iterations of the SIMP algorithm. We save these values for a subset of a dataset (around 13.5K samples) during data generation with an average of 13 iterations per Initial Strain Energy. Following a similar approach as described in \citet{RADE2021104483}, we then train an encoder-decoder style convolutional neural network which takes Initial Strain Energy and given iteration design as input to predict the strain energy of that given iteration's design. 

In \figref{fig:total_kde}, we plot the distribution of predicted strain energy for all 2,000 generated designs against the corresponding 100 SIMP-Optimized designs from the evaluation set. From this figure, we can conclude that, in the aggregate, there is no substantial difference in performance between designs generated by our framework and the SIMP-Optimized designs. Please refer to \figref{fig:kde_subplots} for per-sample plots of strain energy distributions.

Additionally, we evaluate the volume fraction, which is a constraint that specifies the volume of material to be removed from the initial design. Here, we define the volume fraction as the ratio of occupied voxels to unoccupied voxels. The mean absolute error between the average volume fraction across the 20 generated designs and the corresponding SIMP-Optimized design was $0.0989$.

\subsection{Design Translation}
\begin{figure}[h!]
	\centering
    \includegraphics[trim=0 0 0 0,clip,width=0.95\linewidth]{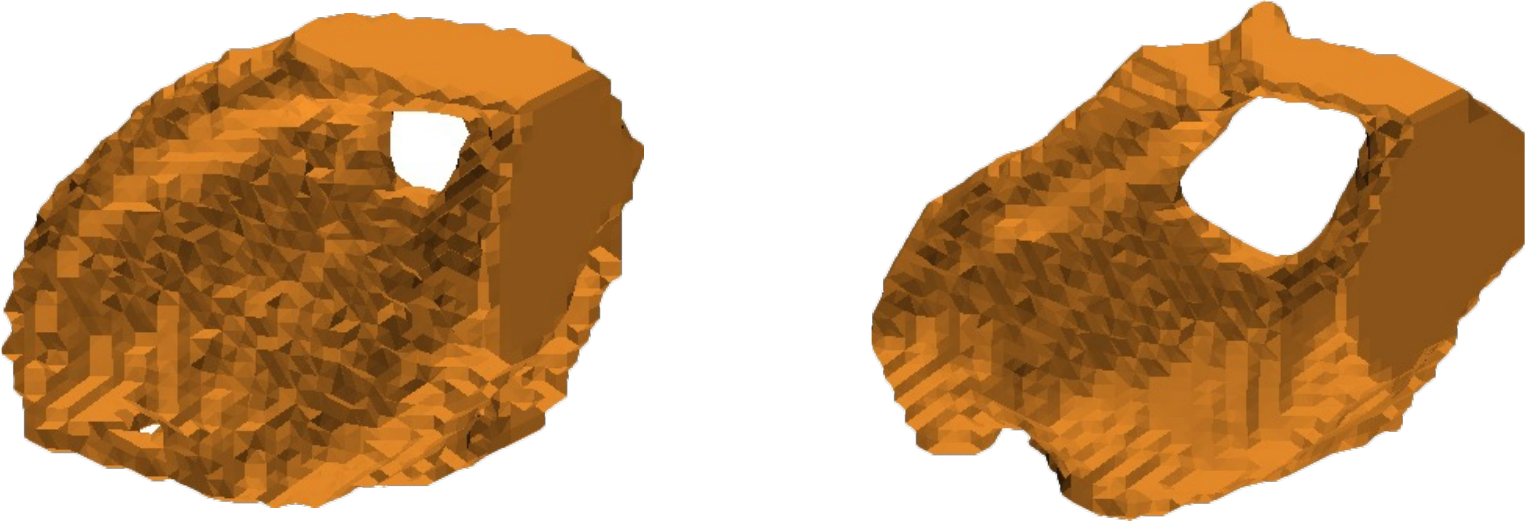}
    \caption{The framework is also flexible enough to edit pre-existing designs. The trained diffusion model will denoise a partially noised pre-existing design latent from the pre-trained SIMP-Optimized Density Encoder. The SIMP-Optimized density is on the left and the LDM-Translated density is on the right.}
	\label{fig:exampleDesignTranslation}
\end{figure}

Our evaluation of the Design Translation task is entirely qualitative, but given that the initial translated designs are SIMP-Optimized, the quantitative results shown for design generation should generalize. \figref{fig:exampleDesignTranslation}, shows one example of design translation and supports our claim of generating a similar yet unique design when given an initial starting point. 

\subsection{Scaling and Compute}
\begin{figure*}[t!]
    \centering
    \includegraphics[width=\linewidth,trim={2.5in, 3.0in, 2.5in, 3.0in},clip]{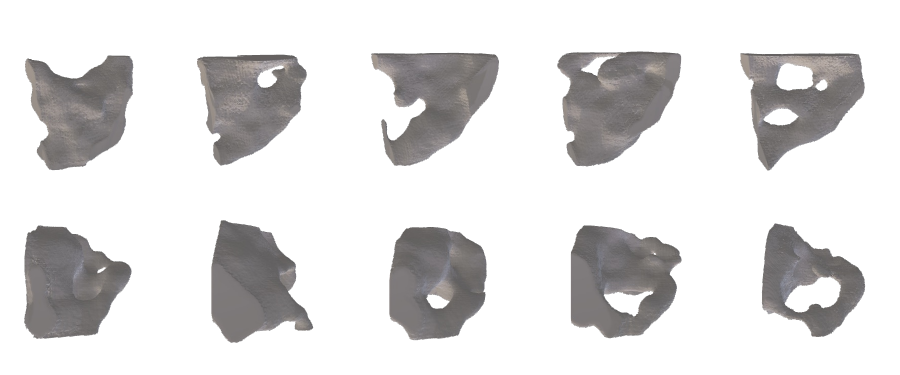}
    \caption{Example generated designs. Each row corresponds to a single Initial Strain Energy sample, and each column is a different generated design. All designs are in the $128^3$ resolution.}
    \label{fig:designs_128cube}
\end{figure*}

Thus far, all analyses described are for generated designs in $32^3$ voxel grids. In \figref{fig:designs_64cube} and \figref{fig:designs_128cube}, we demonstrate the scaling capabilities of our framework for LDM-Generated designs in $64^3$ and $128^3$ voxel grids.

Due to the structure of LDM's, a majority of the learning burden is placed on the external VAE. This means that as long as the VAE achieves high performance on the general reconstruction task, we can confidently train a DM in its latent space. With this in mind, we trained the VAEs for each resolution to have the same size of the latent representation, $[4,8,8,8]$ for each encoder. In doing so, we can use the same model architecture for the internal DM across different resolutions, keeping computational costs in check as we scale.

The VAE training times differ dramatically across resolutions, with $32^3$ taking approximately 2 minutes and 20 seconds per epoch and $64^3$ taking approximately 24 minutes per epoch. The $128^3$ resolution was too large to train from scratch on its own dataset, so we implemented a multi-grid training approach adapted from \citet{rade2023multigrid}. In the multi-grid training approach, we initialize the model architecture for the $128^3$ domain such that its latent representation will be $[4,8,8,8]$ for each encoder. Next, we pretrain this model on the $32^3 dataset$ for ten epochs and continue pretraining for one epoch on the $64^3$ data. Finally, we train for an additional single epoch on the target $128^3$ dataset. Once completed, the model achieves an L1 reconstruction loss of 0.0255 on the validation set in less than two hours of training time. Comparatively, if trained exclusively on the $128^3$ dataset, it takes over 10 hours of training time to achieve a similar level of performance.

Even though we structured each VAE architecture to have the same size of latent representation across different resolutions, the DM training times differ substantially. During training, it takes roughly 6, 22, and 95 milliseconds per sample for $32^3$, $64^3$, and $128^3$ resolutions, respectively. The difference in training time can be attributed to the resolution of each sample and the reduction in batch size to account for the increased memory requirements for increasing resolution.

\section{Conclusions}
\label{Sec:Conclusions}

We developed a Latent Diffusion Model framework for the generative design of structural components. It comprises an external multi-headed Variational Autoencoder and an internal Diffusion Model. An added benefit of this formulation is the ability to edit an arbitrary initial design according to its accompanying initial conditions. Additionally, given the dataset used to train this model was generated using the SIMP topology optimization algorithm, the generated designs are inherently near-optimal. Future directions for subsequent work may involve adding more conditioning mechanisms to the Latent Diffusion Model to guide the generative process. One such example would be directly providing the desired volume fraction of the generated component, i.e., generating components of user-defined size. We hope our framework will be a starting point for using deep learning models for component design during the ideation phase.

\section*{Acknowledgement}
This work was partly supported by the National Science Foundation under grant LEAP-HI 2053760. 

\vfill
\pagebreak

{
\balance
\bibliographystyle{unsrtnat}
\bibliography{Refs}
}

\section*{Supplementary Material}
We provide the compliance distributions for all the designs.

\begin{figure*}[t!]
    \centering
    \includegraphics[width=0.88\linewidth]{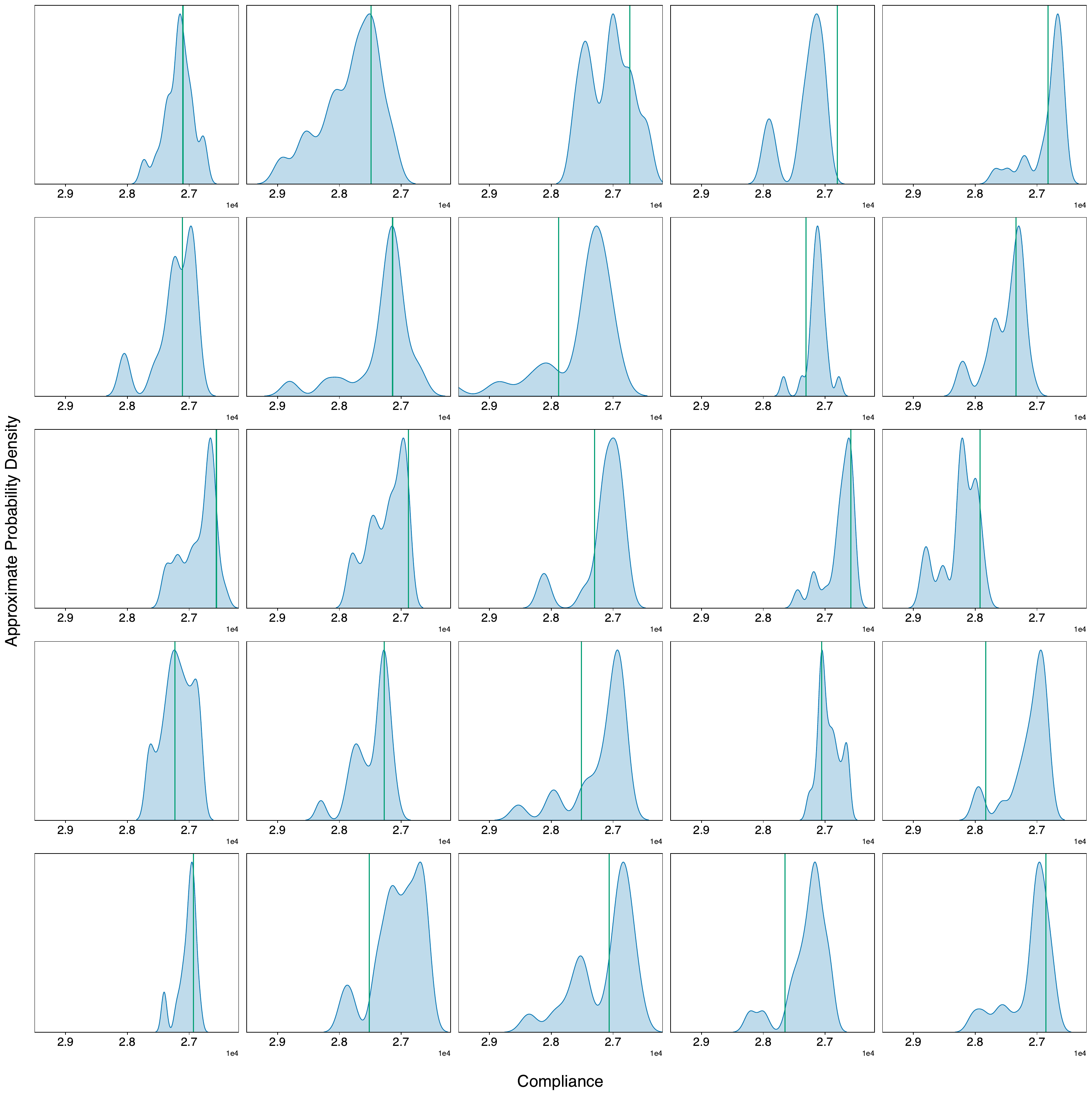}
    \caption{Each subfigure is the predicted total strain energy distribution for 20 generated designs for a given Initial Strain Energy. The red bar is the predicted total strain energy of the corresponding SIMP-Optimized design.}
    \label{fig:kde_subplots1}
\end{figure*}

\begin{figure*}[t!]
    \centering
    \includegraphics[width=0.88\linewidth]{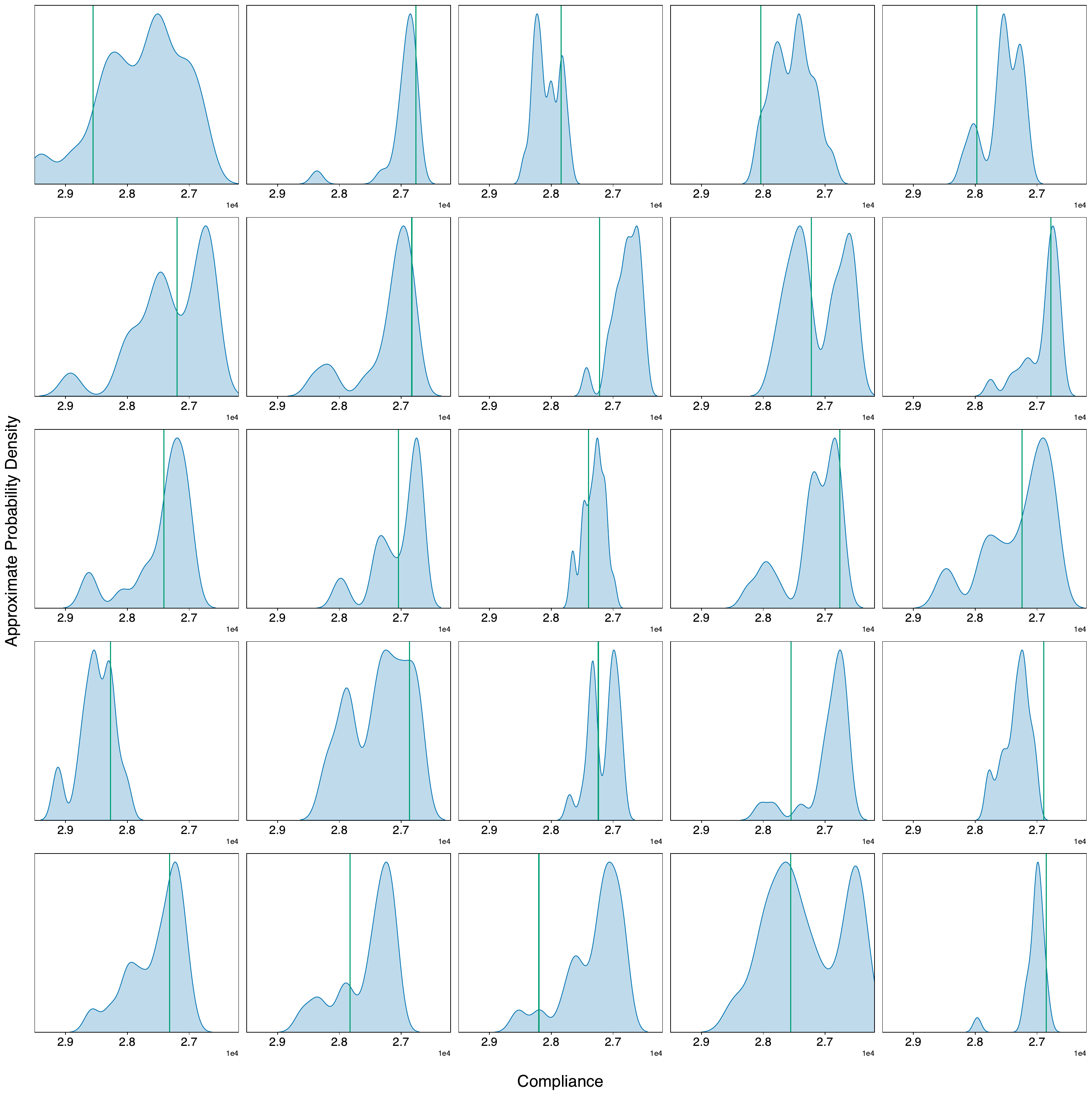}
    \caption{Each subfigure is the predicted total strain energy distribution for 20 generated designs for a given Initial Strain Energy. The red bar is the predicted total strain energy of the corresponding SIMP-Optimized design.}
    \label{fig:kde_subplots2}
\end{figure*}

\begin{figure*}[t!]
    \centering
    \includegraphics[width=0.88\linewidth]{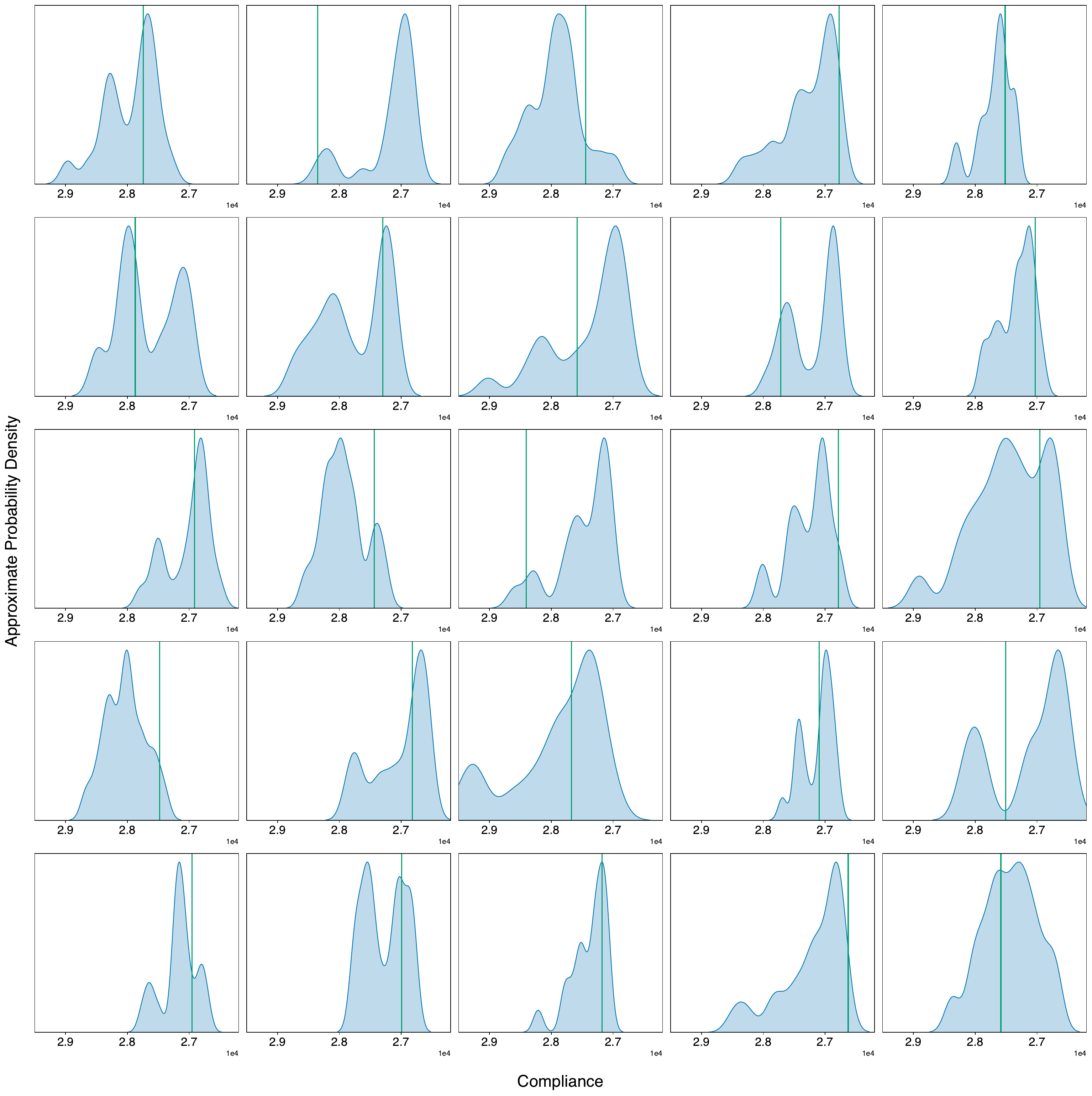}
    \caption{Each subfigure is the predicted total strain energy distribution for 20 generated designs for a given Initial Strain Energy. The red bar is the predicted total strain energy of the corresponding SIMP-Optimized design.}
    \label{fig:kde_subplots3}
\end{figure*}

\begin{figure*}[t!]
    \centering
    \includegraphics[width=0.88\linewidth]{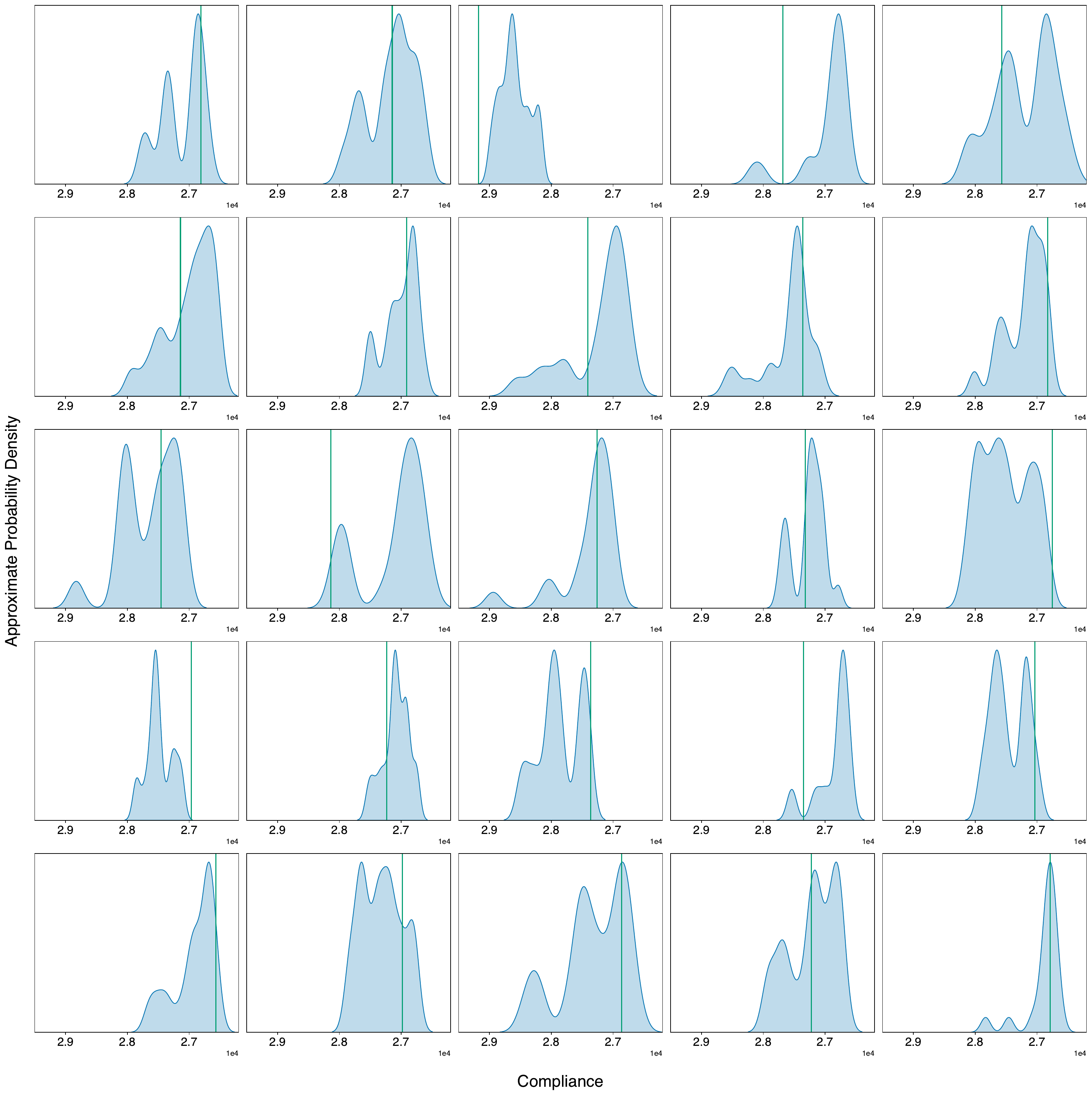}
    \caption{Each subfigure is the predicted total strain energy distribution for 20 generated designs for a given Initial Strain Energy. The red bar is the predicted total strain energy of the corresponding SIMP-Optimized design.}
    \label{fig:kde_subplots4}
\end{figure*}

\end{document}